\documentclass[letterpaper, 10 pt, conference]{ieeeconf}  
\IEEEoverridecommandlockouts
\usepackage{graphics} 
\usepackage{epsfig} 
\usepackage{graphicx} 
\usepackage{amssymb}
\usepackage{subcaption}
\usepackage{url}

\title{\LARGE \bf
Harnessing the Power of Deep Learning Methods in Healthcare: \\ Neonatal Pain Assessment from Crying Sound}

\author{
Md Sirajus Salekin$^{1}$, 
Ghada Zamzmi$^{1*}$,
\thanks{* Ghada Zamzmi current affiliation is National Library of Medicine, National Institutes of Health, MD, USA.}
Rahul Paul$^{1}$,
Dmitry Goldgof$^{1}$, 
Rangachar Kasturi$^{1}$, 
Thao Ho$^{2}$, 
Yu Sun$^{1}$\\
{$^{1}$Department of Computer Science and Engineering, University of South Florida, FL, USA}\\%
{$^{2}$College of Medicine Pediatrics, USF Health, University of South Florida, FL, USA}\\%
}

\begin{document}

\onecolumn
\noindent \huge IEEE Copyright Notice\\

\noindent \normalsize \textcopyright \hspace{0.05cm} 2019 IEEE. Personal use of this material is permitted. Permission from IEEE must be obtained for all other uses, in any current or future media, including reprinting/republishing this material for advertising or promotional purposes, creating new collective works, for resale or redistribution to servers or lists, or reuse of any copyrighted component of this work in other works.\\

\noindent \large {Accepted to be Published in: Proceedings of the 2019 IEEE Healthcare Innovation and Point-Of-Care Technologies Conference (HI-POCT), November 20-22, 2019, MD, USA.}
\twocolumn
\normalsize 

\maketitle
\thispagestyle{empty}
\pagestyle{empty}


\begin{abstract}
Neonatal pain assessment in clinical environments is challenging as it is discontinuous and biased. Facial/body occlusion can occur in such settings due to clinical condition, developmental delays, prone position, or other external factors. In such cases, crying sound can be used to effectively assess neonatal pain. In this paper, we investigate the use of a novel CNN architecture (N-CNN) along with other CNN architectures (VGG16 and ResNet50) for assessing pain from crying sounds of neonates. The experimental results demonstrate that using our novel N-CNN for assessing pain from the sounds of neonates has a strong clinical potential and provides a viable alternative to the current assessment practice.

\end{abstract}


\section{Introduction}
Pain is a universal, yet individual, unpleasant emotional experience that health professionals encounter and deal with in all clinical settings. Although pain is the main reason for people to seek professional healthcare, pain is usually under-/over-recognized and inadequately treated. This is especially true in case of neonates due to the absence of articulation and non-verbal communication (e.g., VAS). The dependence on observers for assessing neonatal pain suffers from different shortcomings. It suffers from the observers' subjectivity as well as their limited ability to continuously observe multiple pain cues (e.g., facial expression and vital signs) at the same time. In addition, detecting and assessing pain of premature neonates is very difficult as the facial and body muscles of these neonates are not well-developed. 

Several studies \cite{zamzmi2016machine,zamzmi2018review,zamzmi2019comprehensive, johnston1993developmental} reported that crying sound is one of the main pain indicators in premature neonates with limited ability to depict facial expression. Further, assessing neonatal pain from crying sound is necessary in cases of occlusion (e.g., prone position or swaddle). The inadequate management of neonatal pain, caused by inaccurate assessment, can lead to serious outcomes \cite{zamzmi2019comprehensive,brummelte2012procedural,zamzmi2018review}. Examples of these outcomes include impaired brain development, altered pain perception, and poorer cognition and motor function. Therefore, developing automated methods that continuously record neonates' sound and use it as a main indicator for accurately assessing pain is crucial to mitigate the shortcomings of the current assessment practice and lead to effective pain management.



Existing automated methods for assessing neonatal pain from crying sound can be broadly divided into: a) handcrafted-based and b) deep learning-based methods. Handcrafted-based methods analyze the sound signal in three different domains, frequency, cepstral, and time Domain.

Pal et al. \cite{pal2006emotion} proposed a frequency domain method, namely Harmonic Product Spectrum (HPS), to extract the Fundamental Frequency ($F0$) method along with the first three formants (i.e., $F1$, $F2$, and $F3$) from crying signals of infants recorded during several states (e.g., pain, hunger, and anger). After extracting the features, k-means algorithm was applied to find the optimal parameters that maximizes the separation between features of different cry types. Combining $F0$, $F1$ and $F2$ produced the best clustering and achieved an accuracy of 91\% for pain. The Cepstral domain of a given signal is created by taking the Inverse Fourier Transform (IFT) of the logarithm of the signal's spectrum. A well-known Cepstral Domain method for analyzing sound is the Mel Frequency Cepstral Coefficients (MFCC). This method extracts useful and representative features (i.e., coefficients) from the sound signal and discard noise. Zamzmi et al. \cite{zamzmi2019comprehensive} calculated MFCC coefficients and LPCC coefficients (Linear Prediction Cepstral Coefficients) from sound segments. The coefficients were extracted using 32 ms Hamming window with 16ms overlapping for LPCC coefficients and 30 ms Hamming window with 10 ms shift MFCC coefficients. The accuracy of assessing neonatal pain using the features extracted from the crying sounds of 31 neonates was 82.35\%.

Vempada et al. \cite{vempada2012characterization} extracted a time domain feature, called Short-time Frame Energies (STE), along with other features to represent spectral and prosodic information. The extracted features were used with Support Vector Machine (SVM) to classify the infant cry into pain and other categories. The proposed method achieved 74.07\% and 80.56\% accuracies using feature and score level fusion, respectively. Instead of using only time or frequency features, Chen et al. \cite{chen2017neonatal} combines the features from both Time and Frequency domain. They extracted 15 features and used Sequential Forward Floating Selection (SFFS) algorithm followed by Directed Acyclic Graph Support Vector Machine (DAG-SVM) algorithm. They have reported an accuracy of 95.45\%, 76.81\%, and 86.36\% for pain, sleepiness, and hunger respectively. Recently, deep learning-based method \cite{salekin2019multi} has gained much popularity as it achieves excellent performance that outperforms the performance of human experts. Despite the popularity of deep learning networks, we are not aware of any work that harnesses the power of these networks for assessing neonatal pain from crying sound. In a different application, deep networks achieved excellent performance when used to analyze the sounds of newborns \cite{severini2019automatic}.

In this paper, we propose a fully automated deep learning-based method for assessing neonatal pain from crying sound. Specifically, we assess neonatal pain using a novel CNN, called Neonatal Convolutional neural network (N-CNN) \cite{zamzmi2019convolutional}, with the spectrogram of neonates' audio signals. We demonstrate the performance of N-CNN by comparing it with the performance of other well-known CNNs such as VGG \cite{simonyan2014very} and ResNet \cite{he2016deep}. As far as we are aware, we are the first to fully investigate the use of deep learning-based methods for assessing pain from crying sounds of neonates recorded in a NICU setting with different background noise (e.g., nurses sounds, equipment sounds, and crying sounds of other neonates). We evaluated all the networks using NPAD \cite{zamzmi2019comprehensive} dataset, which was collected in a real NICU clinical environment (Section IV.A).

\section{Background}
\subsection{Analysis of Visual Representation: Spectrogram Image}
Although using a clean raw signal is the most straightforward method for analyzing sounds, most audio signals recorded in real-world settings are not clean and contain various levels of background noise. Several advanced methods were proposed to reduce background noise \cite{hannun2014deep}. However, the high computational complexity of these methods makes it unsuitable for real-time applications. Empirical experiments have suggested \cite{dennis2014analysis} that spectrogram images can suppress the signal noise while keeping the details of the energy distribution. In addition, studies \cite{dennis2014analysis} have reported that classifying sounds events using a visual representation (spectrogram images) can be more accurate as the energy of sound events are concentrated in a small number of spectral components with a distinctive time-frequency representation \cite{dennis2014analysis}. Further, using a visual representation allows to harness the state-of-the-art deep learning methods as the majority of these methods are designed for image classification (e.g., VGG). 

\begin{figure}[t]
    \centering
    \begin{subfigure}[b]{0.22\textwidth}
        \includegraphics[width=\textwidth]{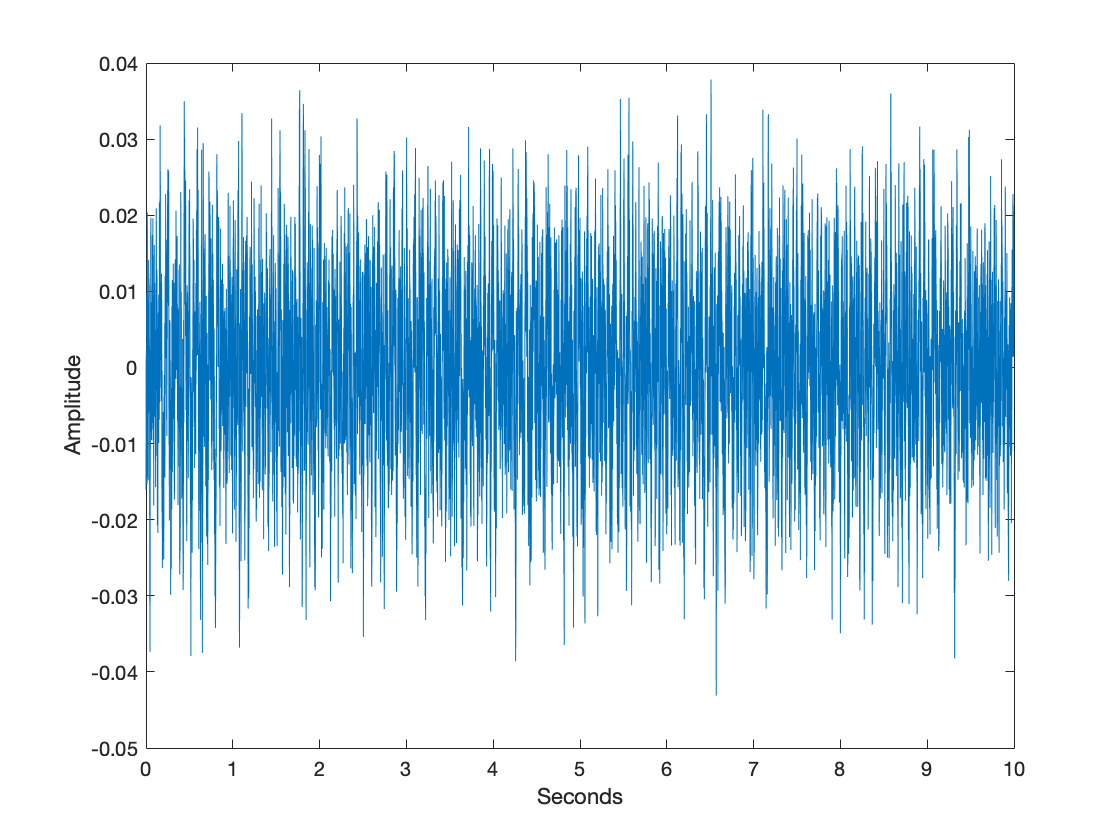}
        \caption{No-pain signal}
        \label{fig:nopain_original}
    \end{subfigure}
    \quad
    \begin{subfigure}[b]{0.22\textwidth}
        \includegraphics[width=\textwidth]{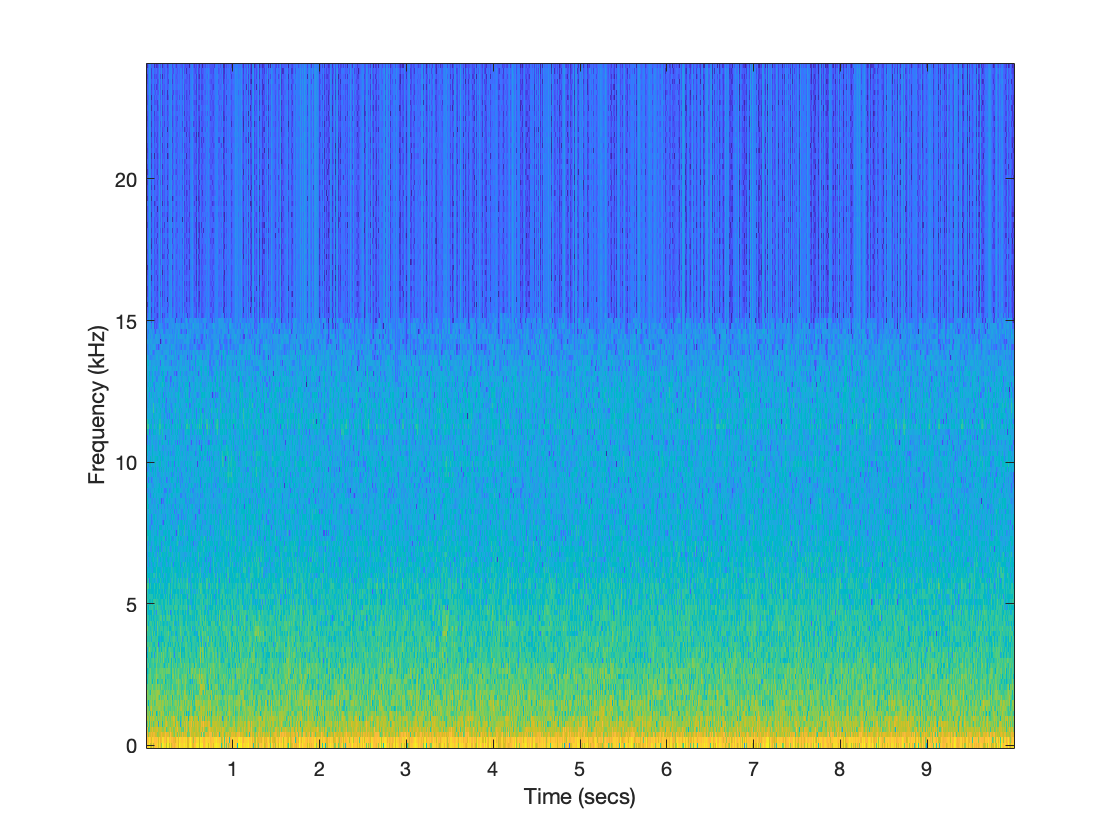}
        \caption{Spectrogram image}
        \label{fig:nopain_spectrogram}
    \end{subfigure}
    
 \caption{Sample of original no-pain audio signal and its corresponding spectrogram image}
 \label{fig_nopain_original_spectrogram}
\end{figure}

\begin{figure}[t]
    \centering
    \begin{subfigure}[b]{0.22\textwidth}
        \includegraphics[width=\textwidth]{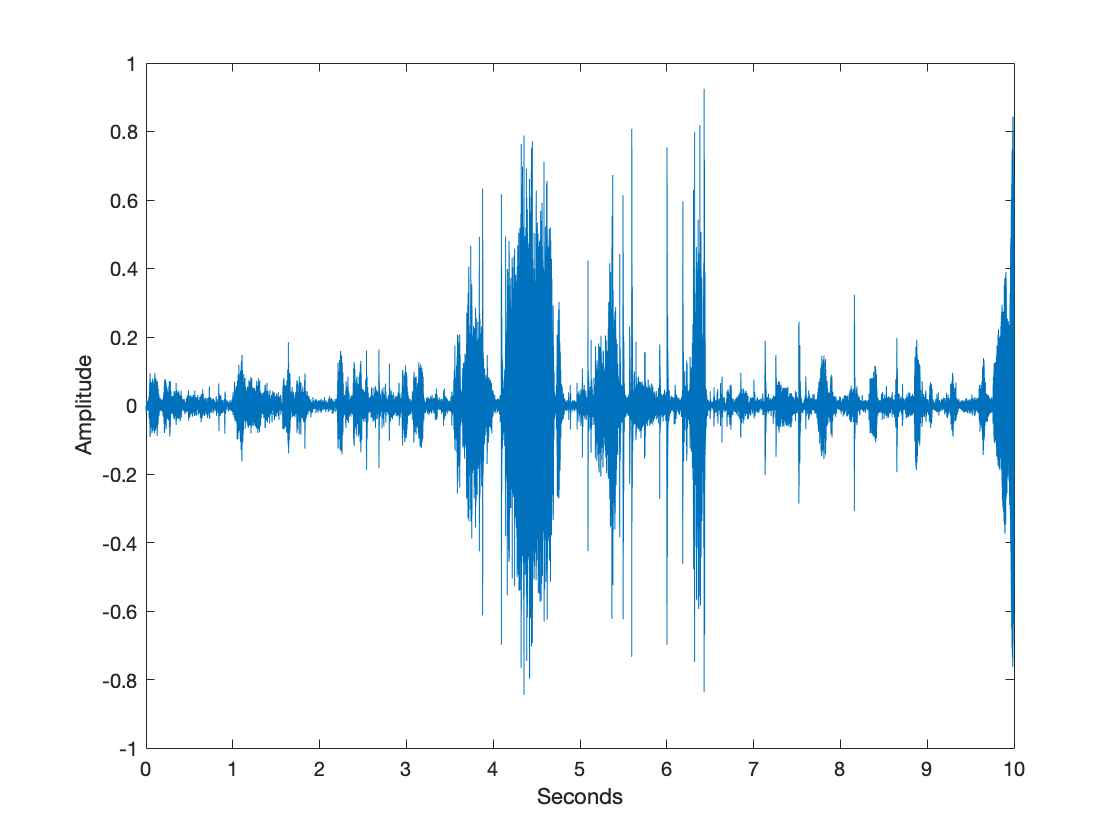}
        \caption{Pain signal}
        \label{fig:pain_original}
    \end{subfigure}
     \quad
    \begin{subfigure}[b]{0.22\textwidth}
        \includegraphics[width=\textwidth]{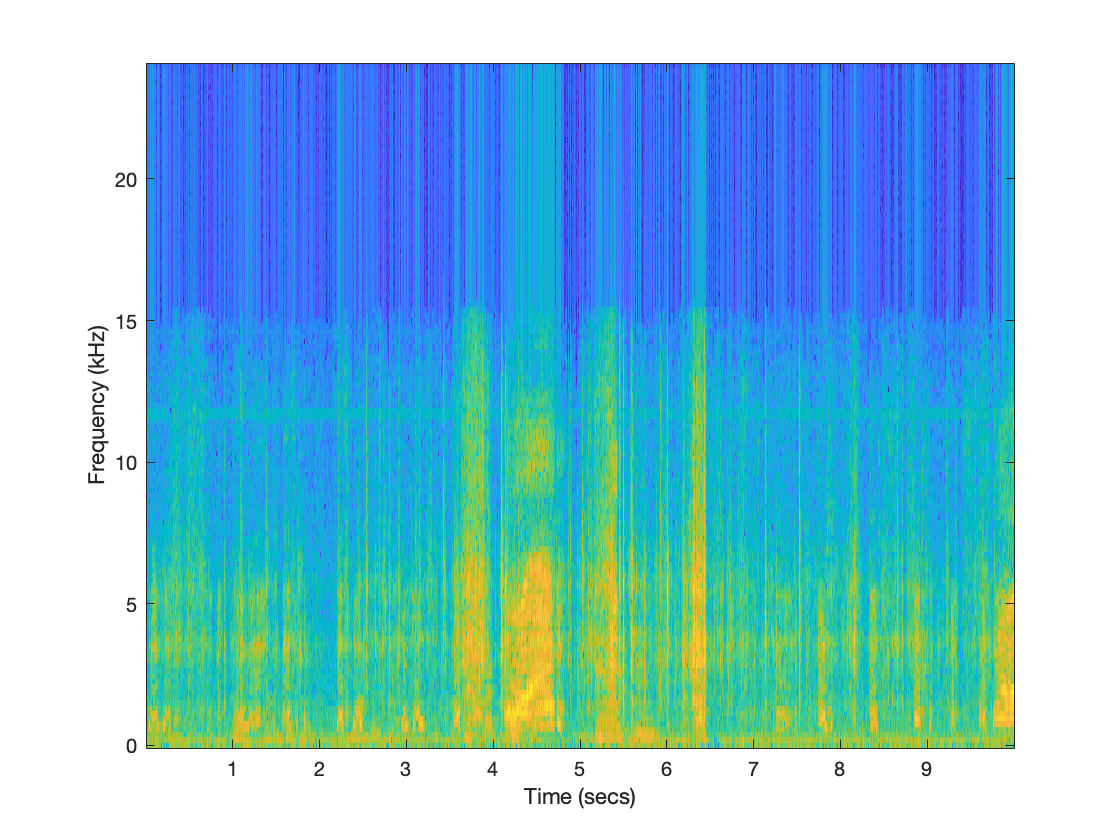}
        \caption{Spectrogram image}
        \label{fig:pain_spectrogram}
    \end{subfigure} 
 
 \caption{Sample of original pain audio signal and its corresponding spectrogram image}
 \label{fig_pain_original_spectrogram}
\end{figure}

Spectrogram image \cite{oppenheim1999discrete} provides a visual representation of the audio signal. As shown in Figure \ref{fig_nopain_original_spectrogram} and Figure \ref{fig_pain_original_spectrogram}, a spectrogram image provides a 2-D or 3-D visual representation of change for every frequency component (y-axis) of an audio signal with respect to time (x-axis). The higher energy is represented by a brighter pixel compared to the lower energy. Since the spectrogram image shows the change of every frequency component over time, the noise becomes easily identifiable. 

\subsection{N-CNN}
Neonatal Convolutional Neural Networks (N-CNN) \cite{zamzmi2019convolutional} is a lightweight cascaded CNN developed specifically for neonatal population. It has three branches where each branch extracts different types of features followed by combining the features of all branches to generate a robust image feature vector \cite{zamzmi2019convolutional}. The left branch includes a pooling layer using a $10\times 10$ filter. The right branch includes a convolutional layer of 64 filters with a $5\times 5$ convolution followed by a similar pooling layer like the left branch. The central branch consists of two consecutive versions of similar type of convolutional layers and pooling layers. Finally, all the layers are merged and followed by a convolutional-pooling layer. The entire network combines image-specific features with edges and blobs and generates a robust image feature description. Before the final classification layer, an L2 regularizer and a dropout were applied. N-CNN uses RMSprop for the gradient descent optimizer with a learning rate of 0.0001. N-CNN has shown promising performance in case of facial expression analysis as reported in \cite{zamzmi2019convolutional}.


\section{Methodology}
\subsection{Preprocessing}
In our experiments, we extracted audio events from all videos of 31 subjects. Each of the extracted audio starts immediately before the painful procedure and ends after the completion of the painful procedure. We converted these audio segments into spectrogram images of size $120\times 120$ and used these images as input for N-CNN. Note that we experimented with different image sizes and used the one ($120\times 120$) since it achieves the best performance. As for VGG16 and ResNet50, we used an image size of $224\times 224$. 

\subsection{Signal Augmentation}
Because training CNNs requires a large amount of data, we performed data augmentation in the audio segments (pain and no-pain events). Each audio segment was augmented by adding three frequencies ($f/2$, $f/3$, $2f/3$), six different levels of noise ($0.01$, $0.05$, $0.001$, $0.005$, $0.003$, and $0.03$), and a combination of both frequency and noise (for example, $f/2$ with noise $0.01$ or $f/3$ with noise $0.001$). Thus, a total of 27 augmented signals was generated for each audio event (3 ways for frequency change, 6 levels of Noise, 18 ways for different combinations of both frequency and noise); 4914 ($182 \times 27 $) augmented signals. Figure 3 and Figure \ref{fig_original_sepctogram_aug} show examples of augmented spectrogram images for no-pain and pain events, respectively. 


\begin{figure}[t]
    \centering
    \begin{subfigure}[b]{0.22\textwidth}
        \includegraphics[width=\textwidth]{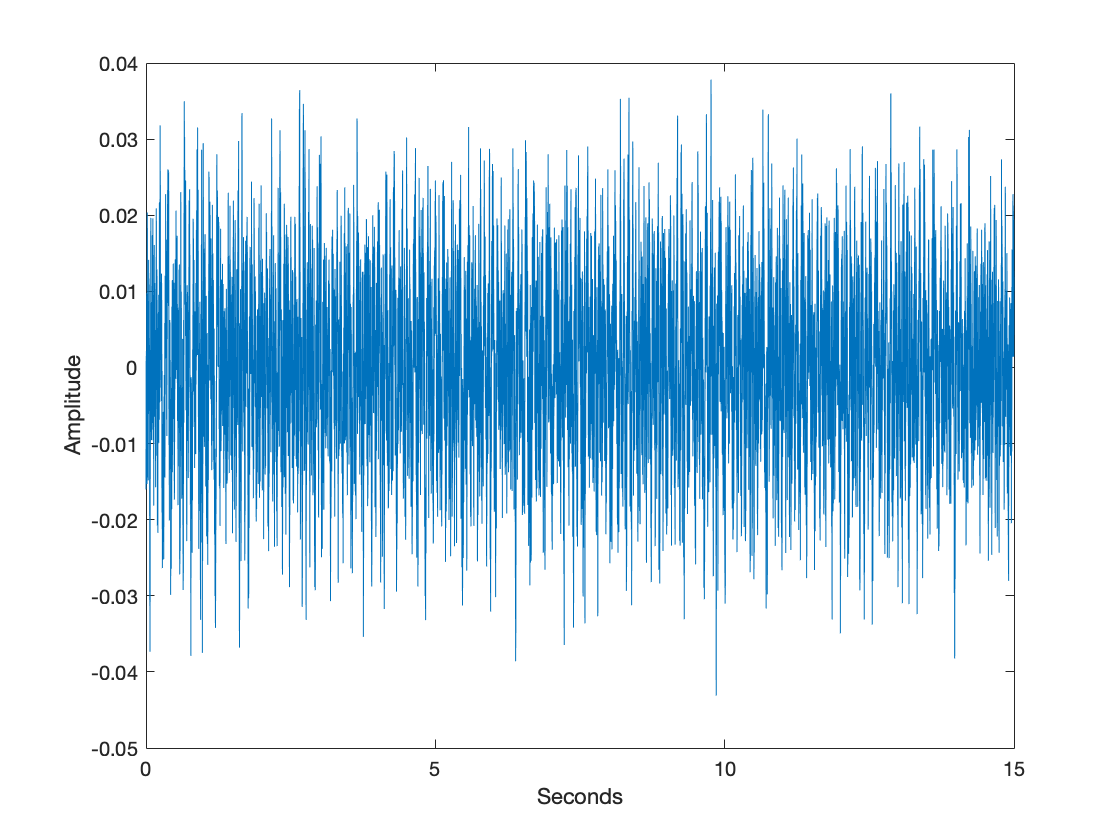}
        \caption{No-pain signal ($2f/3$)}
        \label{fig:nopain_original_aug}
    \end{subfigure}
    \quad
    \begin{subfigure}[b]{0.22\textwidth}
        \includegraphics[width=\textwidth]{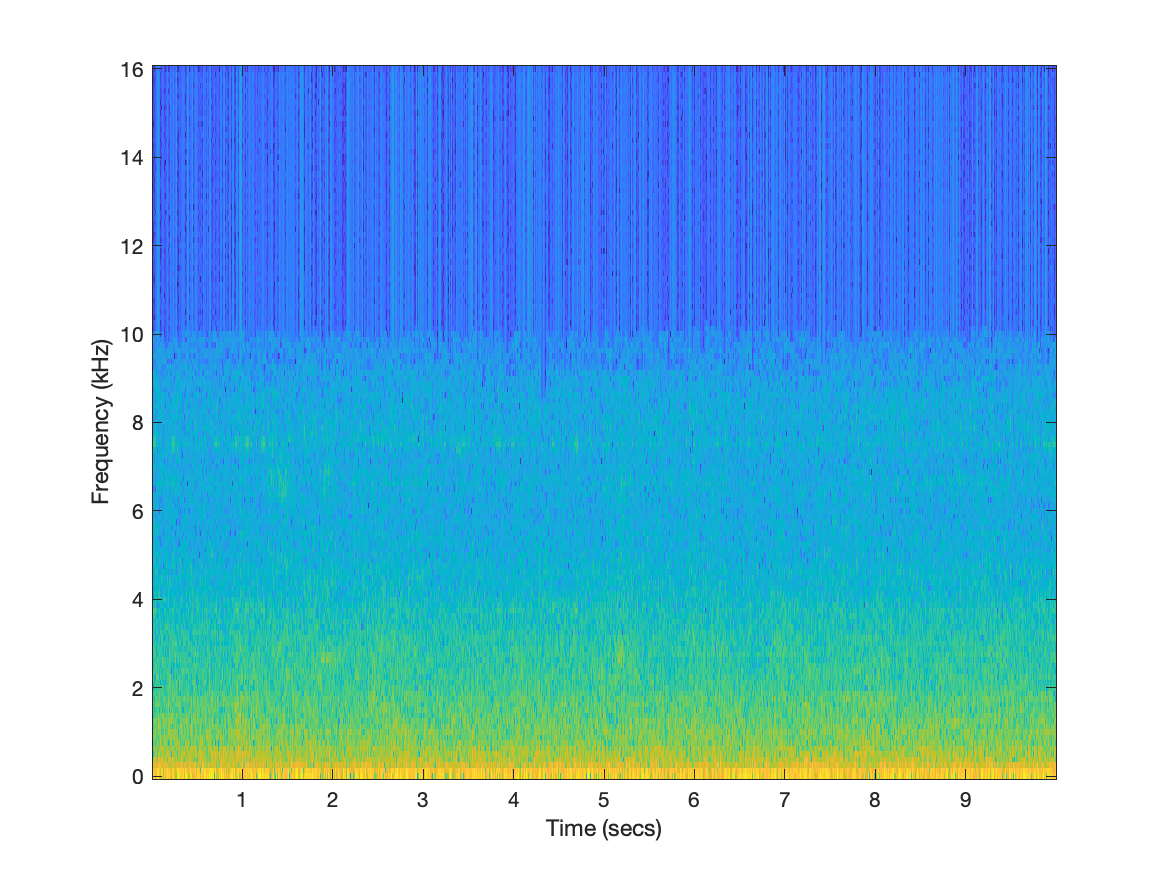}
        \caption{Spectrogram image}
        \label{fig:nopain_spectrogram_aug}
    \end{subfigure}
 
 \caption{Sample of augmented image of the original no-pain audio signal and its corresponding spectrogram image}
 \label{fig_original_sepctogram_aug}    
\end{figure}
\begin{figure}[t]
    \centering
     \begin{subfigure}[b]{0.22\textwidth}
        \includegraphics[width=\textwidth]{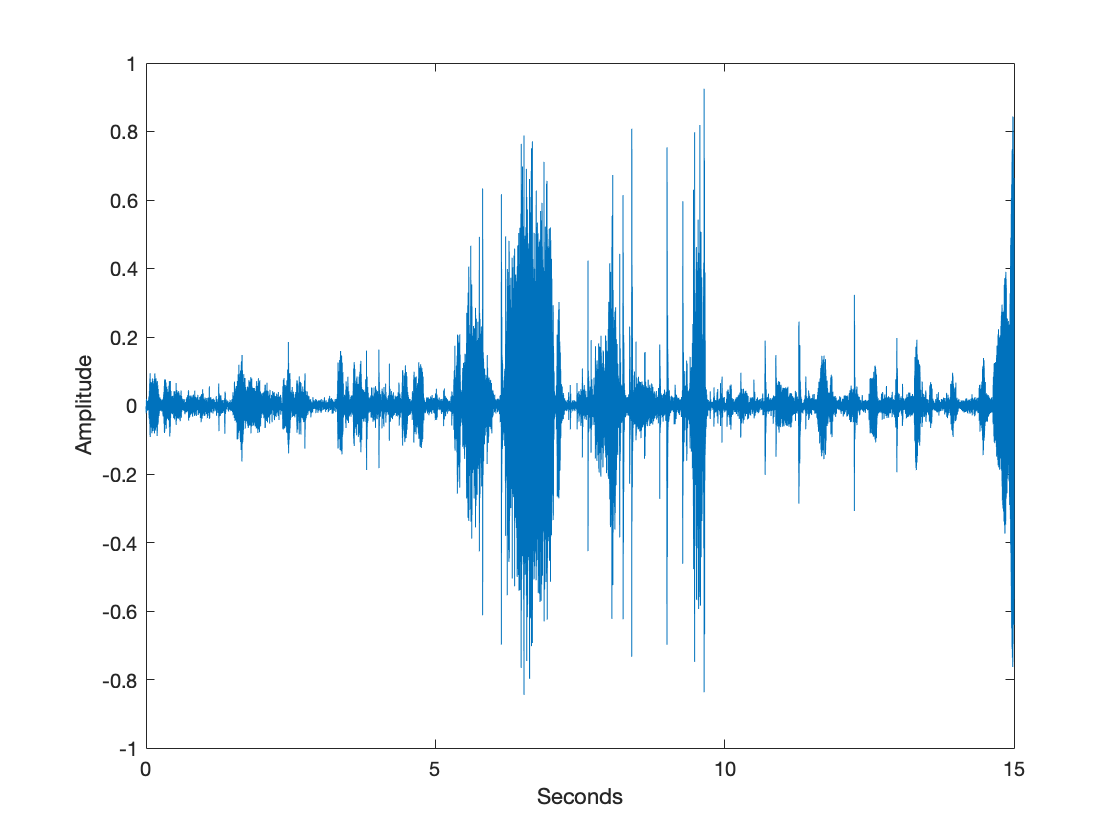}
        \caption{Pain signal ($2f/3$)}
        \label{fig:pain_original_aug}
    \end{subfigure}
     \quad
    \begin{subfigure}[b]{0.22\textwidth}
        \includegraphics[width=\textwidth]{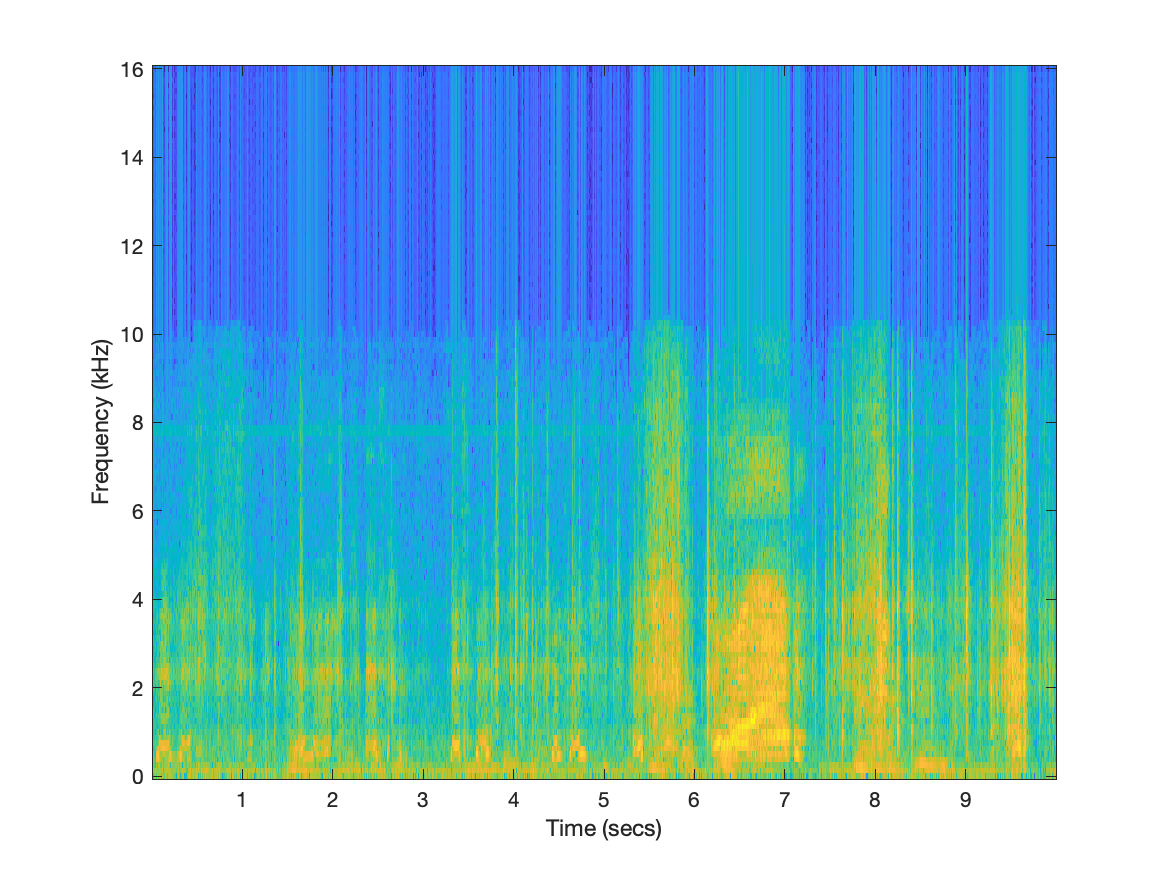}
        \caption{Spectrogram image}
        \label{fig:pain_spectrogram_aug}
    \end{subfigure}   
 
 \caption{Sample of augmented image of the original pain audio signal and its corresponding spectrogram image}
 \label{fig_original_sepctogram_aug}
\end{figure}


\subsection{Deep Learning Based Approach}
Recently, different deep learning architectures such as VGG16 \cite{simonyan2014very}, ResNet50 \cite{he2016deep}, N-CNN \cite{zamzmi2019convolutional} are applied to identify neonatal pain from facial expression and showed promising performance. In this paper, we investigated the use of our novel N-CNN along with two well-known CNNs architectures (e.g., VGG16 and ResNet50) for neonatal pain assessment. Since deep learning approaches work well with images and spectrogram images can easily suppress the noise of audio signals, we applied these architectures on the spectrogram images of the audio signals. We trained the novel N-CNN, VGG16, and ResNet50 using the augmented dataset (Section III.B). The complete training parameters and the network architecture of N-CNN can be found in \cite{zamzmi2019convolutional}. We fined tuned VGG16 \cite{simonyan2014very} and ResNet50 \cite{he2016deep}, originally pre-trained using ImageNet, as summarized in Table \ref{tab_vgg16_resnet50_finetuned}.


\begin{table}
    \begin{center}
    \caption{Tuned VGG16 and ResNet50 architectures} 
    \begin{tabular}{|c|c|}
    
    \hline
    \multicolumn{2}{|c|}{\textbf{VGG16 Architecture}} \\
    \hline
    
    Conv3 & $64 \times 3 \times 3$, st. 1, pad 1 \\
    Conv 1-2 & $64 \times 3 \times 3$, st. 1, pad 1 \\
    \hline
    Conv 2-1 & $128 \times 3 \times 3$, st. 1, pad 1 \\
    Conv 2-2 & $128 \times 3 \times 3$, st. 1, pad 1 \\
    \hline
    Conv 3-1 & $256 \times 3 \times 3$, st. 1, pad 1 \\
    Conv 3-2 & $256 \times 3 \times 3$, st. 1, pad 1 \\
    Conv 3-3 & $256 \times 3 \times 3$, st. 1, pad 1 \\
    \hline
    Conv 4-1 & $512 \times 3 \times 3$, st. 1, pad 1 \\
    Conv 4-2 & $512 \times 3 \times 3$, st. 1, pad 1 \\
    Conv 4-3 & $512 \times 3 \times 3$, st. 1, pad 1 \\
    \hline
    Conv 5-1 & $512 \times 3 \times 3$, st. 1, pad 1 \\
    Conv 5-2 & $512 \times 3 \times 3$, st. 1, pad 1 \\
    Conv 5-3 & $512 \times 3 \times 3$, st. 1, pad 1 \\
    \hline
    Full 6 & 512 dropout =0.5 , relu \\
    \hline
    Full 7 & 512 dropout =0.5, relu\\
    \hline
    Full 8 & 1, sigmoid \\
    \hline
    
    \hline
    \multicolumn{2}{|c|}{\textbf{ResNet50 Architecture}} \\
    \hline
    
    Global Average Pooling & Base model output\\
    \hline
    Dropout &  0.5\\
    \hline
    Full 1 & 1, sigmoid \\
    \hline
    \end{tabular}
    \label{tab_vgg16_resnet50_finetuned}
    \end{center}
\end{table}


\section{Experimental Results and Discussion}
\subsection{Neonatal Pain Assessment Database (NPAD)}
We collected vocal, visual, and vital signs data from 31 neonates (50\% female) as part of approved USF study (IRB Pro00014318). The neonates age ranges from 32 0/7 to 40 0/7 (mean: 35.9) gestational weeks. The visual data, which includes face, head, and body, was collected using a GoPro Hero 5 black camera. The vocal data, which consists of the neonates sounds and background noise (e.g., nurse/equipment sounds), was recorded using the microphone of the same camera. To mark the ground truth events (pain assessment scores) given by trained nurses, we used a clapperboard. 

We collected data from neonates while at baseline, during procedural painful procedures (e.g. heel lancing), and after the completion of the painful procedure (i.e., recovery). All procedures were performed as clinically indicated procedures; no procedures were performed for study purposes. To get the ground truth labels, trained nurses were asked to assess pain using NIPS (Neonatal Infant Pain Scale) \cite{hudson2002validation} scoring tool. The nurses observed the neonate and provided NIPS scores prior to the procedure (baseline), during, and after the completion of the procedure. Detailed description of NPAD dataset can be found in \cite{zamzmi2019comprehensive}. 

\subsection{Evaluation Protocol}
In our experiments, we used two evaluation protocols. Deep learning methods require a larger dataset to train the model, but in clinical practice, smaller dataset is relatively common. Therefore, we used leave-one-subject-out (LOSO) cross validation evaluation protocol with deep learning-based methods. In addition to LOSO, we used 10 folds cross validation when we compare the performance of the proposed N-CNN with existing handcrafted-based methods as these methods were evaluated using subject-based 10 folds cross validation. 

\subsection{Pain Assessment From Crying Sound}
The experimental results of using deep learning-based methods for assessing neonatal pain from sounds are promising. Table \ref{tab_deeplearning} shows the pain assessment performance of our N-CNN, VGG16 \cite{simonyan2014very}, and ResNet50 \cite{he2016deep} architectures. Both VGG16 \cite{simonyan2014very} and N-CNN achieved similar performance with an accuracy of 96.77\% and AUC of 0.94. ResNet50 \cite{he2016deep} achieved 83.87\% accuracy and 0.83 AUC. Note that the performance of the proposed N-CNN is comparable to the fine-tuned VGG16 and higher than the fine tuned ResNet50 although N-CNN has much smaller training parameters (2nd column of Table \ref{tab_deeplearning}). 

We also compared the assessment performance of N-CNN network with other existing methods \cite{pal2006emotion,petroni1995identification,zamzmi2019comprehensive}. As shown in the last row of Table III, using our novel N-CNN with spectrogram images achieved the highest performance (91.20\% accuracy and 0.94 AUC). The second highest assessment performance (82.35\% accuracy and 0.69 AUC) was obtained using LPCC-MFCC feature method \cite{zamzmi2019convolutional}. As for the other two methods, we included their performance as reported in the papers. Using MFCC features with NN \cite{petroni1995identification} for detecting neonatal pain cry achieved an accuracy of 78.56\% while using fundamental frequency features with K-mean \cite{pal2006emotion} achieved an accuracy of 74.21\%. Note that AUC metric was not reported for these methods. In addition, these two methods (\cite{petroni1995identification} and \cite{pal2006emotion}) were evaluated using different datasets. N-CNN and \cite{zamzmi2019comprehensive} was evaluated using NPAD \cite{zamzmi2019comprehensive}. 


\begin{table}[t]
\centering
\caption{Neonatal pain assessment
from sound (spectrogram images) using deep learning and LOSO protocol}
\begin{tabular}{|l|c|c|c|}
\hline
    Approach  & Total parameters & Accuracy (\%) & AUC  \\ \hline
    VGG16 & 27,823,425 & 96.77 & 0.94\\ \hline
    ResNet50 & 23,688,065  &  83.87 & 0.83 \\ \hline
    N-CNN  & 72593 & 96.77 & 0.94 \\ \hline
\end{tabular}
\label{tab_deeplearning}
\end{table}


\begin{table}[t]
\centering
\caption{N-CNN and handcrafted methods, 10 folds CV}
\begin{tabular}{|l|c|c|}
\hline
    Approach  & Accuracy (\%) & AUC    \\ \hline
    MFCC + NN \cite{petroni1995identification} & 78.56 & - \\ \hline
    Fundamental Frequency + K-mean \cite{pal2006emotion} & 74.21 & - \\ \hline
    LPCC/MFCC+SVM \cite{zamzmi2019comprehensive} & 82.35 &  0.69 \\ \hline
    Spectrogram + N-CNN (Proposed) & 91.20 & 0.91  \\ \hline
\end{tabular}
\label{tab_comparision}
\end{table}


\section{Conclusion and Future Directions}
Neonatal pain assessment in clinical environments is challenging due to its intermittent nature as well as observer subjectivity and limited ability to accurately monitor multiple pain cues. Facial/body occlusion can be common in NICU settings due to clinical condition (e.g., oxygen mask), prone position, or other factors. Hence, automated systems that assess pain based on analysis of sounds can provide a reliable and effective alternative to the current practice. In this paper, we fully investigate the use of deep learning-based methods for neonatal pain assessment. Specifically, we used a proposed N-CNN and two well-known CNNs for analyzing spectrogram images of neonates' sounds recorded while at baseline and experiencing painful procedures. These results are promising and prove the feasibility of automatic neonatal pain assessment based on analysis of crying sounds. In the future, we plan to evaluate the proposed approach on a larger dataset of infants recorded during both procedural and postoperative pain. We also plan to expand our approach to include different levels of pain as well as different states (e.g., hunger). Finally, we plan to build a neonatal neural network that combines crying sound with facial expression and body movement to create a multimodal assessment of neonatal pain.

\section*{ACKNOWLEDGMENT}
We are grateful for the parents who allowed their babies to take part in this study and the entire neonatal staff at Tampa General Hospital for their help and cooperation in the data collection. 


\bibliographystyle{IEEEtran}
\bibliography{bibliography.bib}

\end{document}